\documentclass[11pt]{article}

\usepackage[a4paper,margin=1in]{geometry}
\usepackage[T1]{fontenc}
\usepackage[utf8]{inputenc}
\usepackage{lmodern}
\usepackage{microtype}
\usepackage{indentfirst}
\usepackage{amsmath,amssymb}
\usepackage{graphicx}
\usepackage{booktabs}
\usepackage{multirow}
\usepackage{array}
\usepackage{tabularx}
\usepackage{enumitem}
\usepackage{caption}
\usepackage{subcaption}
\usepackage{cite}
\usepackage{url}
\usepackage{xcolor}
\usepackage{hyperref}
\usepackage{tikz}

\usepackage[T1]{fontenc}
\usepackage{lmodern}
\usepackage{xcolor}
\usetikzlibrary{arrows.meta,calc,positioning,shapes.geometric}

\definecolor{panelblue}{HTML}{4D6FAE}
\definecolor{frontblue}{HTML}{4D94CF}
\definecolor{topblue}{HTML}{77B0DA}
\definecolor{sideblue}{HTML}{2C719E}
\definecolor{orange}{HTML}{FF6B2C}
\definecolor{darkorange}{HTML}{C54B08}
\definecolor{green}{HTML}{00B77D}
\definecolor{lightgreen}{HTML}{9BCB75}
\definecolor{yellowatt}{HTML}{FFF500}
\definecolor{convfront}{HTML}{A54E49}
\definecolor{convside}{HTML}{CF5B58}
\definecolor{convtop}{HTML}{CC7771}
\definecolor{spatialfront}{HTML}{568692}
\definecolor{spatialside}{HTML}{3D6F78}
\definecolor{spatialtop}{HTML}{7FA4AC}
\definecolor{poolblue}{HTML}{777DC7}
\definecolor{poolgreen}{HTML}{91AE58}
\usetikzlibrary{arrows.meta,positioning,fit,calc,shapes.multipart}

\hypersetup{
  colorlinks=true,
  linkcolor=blue!50!black,
  citecolor=blue!50!black,
  urlcolor=blue!50!black
}

\setlist[itemize]{leftmargin=2em,itemsep=0.25em,topsep=0.35em}
\title{An Intelligent Cloud--Edge Multimodal Interaction System for Robots}
\author{
Zihan Guo\\ 
Hainan University\\
Haikou, China\\ 
\texttt{zihan.guo@hainanu.edu.cn}
\and
Xiaoqi Li\\ 
Hainan University\\Haikou, China\\ 
\texttt{csxqli@hainanu.edu.cn}
}
\date{}
\begin{document}
\maketitle  

\maketitle

\begin{abstract}
Robust human--robot interaction in complex environments requires accurate gesture perception, semantic scene understanding, and reliable task planning under limited onboard computing resources. This paper presents a cloud--edge multimodal interaction framework that integrates an enhanced YOLO-based gesture detector with coordinated large language model (LLM) and vision--language model (VLM) agents. The proposed detector, incorporates the Convolutional Block Attention Module (CBAM) into the neck and replaces the baseline bounding-box regression objective with Distance-IoU (DIoU) loss. These modifications improve feature discrimination and localization for small or partially occluded gestures in complex backgrounds. The cloud layer performs gesture detection, scene understanding, multimodal fusion, and action planning, whereas the TonyPi robot locally handles data acquisition, communication, action execution, and feedback. Experiments on a public gesture dataset and a custom dataset show that YOLO-DC achieves precision values of 98.9\% and 95.0\%, with mAP@0.5 values of 90.7\% and 92.7\%, respectively. System-level evaluation yields success rates of 95\%, 88\%, and 82\% for single-action, composite-action, and vision-dependent tasks. A 30-participant evaluation yields an overall mean satisfaction score of 3.69 out of 5. These results demonstrate the feasibility of combining refined gesture detection with multimodal agents for resource-constrained robotic interaction.
\end{abstract}


\section{Introduction}
Recent advances in human--robot interaction have accelerated the transition from conventional button-based interfaces to multimodal systems that combine speech, vision, and gestures. Gesture recognition provides a natural and intuitive communication channel and has broad potential in service robotics, assistive systems, and smart homes. Nevertheless, reliable gesture-based interaction in dynamic environments remains difficult because lightweight detectors may lose accuracy for small or occluded gestures, while resource-constrained robots often lack the semantic reasoning and task-planning capabilities required for complex instructions.

Deep-learning-based object detectors, particularly the YOLO family \cite{ref1,ref2,ref3,ref4,ref5,ref6,ref7,ref8}, have substantially advanced real-time gesture recognition. Other representative architectures, including Faster R-CNN \cite{ref9}, SSD \cite{ref10}, and DETR-based models \cite{ref12,ref13}, also provide strong detection performance. However, illumination variations, background clutter, partial occlusion, and diverse gesture poses continue to degrade accuracy. Attention mechanisms \cite{ref15,ref16,ref17,ref18,ref19} and improved bounding-box regression losses \cite{ref20,ref21,ref22,ref23,ref24} have therefore been explored to strengthen feature representation and localization.

At the system level, traditional robots frequently rely on a single input modality and fixed command mappings, which limit their ability to interpret context and decompose complex tasks. Vision--language models (VLMs) and large language models (LLMs) \cite{ref25,ref26,ref27,ref28,ref30} offer new opportunities for multimodal reasoning. Models such as CLIP \cite{ref31}, BLIP \cite{ref32,ref33}, LLaVA \cite{ref34,ref35}, Flamingo \cite{ref36}, and MiniGPT-4 \cite{ref37} have demonstrated strong visual--linguistic understanding. Robotics frameworks such as SayCan \cite{ref40}, VoxPoser \cite{ref41}, RT-2 \cite{ref42}, and CLIPort \cite{ref43} further illustrate how language-conditioned reasoning can support action generation. Directly deploying such computationally demanding models on platforms such as Raspberry Pi or TonyPi, however, remains challenging \cite{ref52,ref53}.

To address these limitations, this study develops a unified cloud--edge interaction framework that combines refined gesture perception with coordinated LLM and VLM agents. The main contributions are as follows:
\begin{itemize}
  \item \textbf{Refined gesture detection.} We construct YOLO-DC by inserting CBAM into the neck of YOLO11n and using DIoU loss for bounding-box regression. Experiments on the public and custom datasets show improved precision and mAP@0.5 in scenarios involving small targets and complex backgrounds.
  \item \textbf{Robot-oriented multimodal agent architecture.} We design a dual-agent architecture that integrates gesture detection, visual question answering, language understanding, rule-based validation, and finite-state-machine control. This architecture supports cross-modal interpretation and structured action planning.
  \item \textbf{Cloud--edge collaborative deployment.} Computationally intensive inference and reasoning are performed in the cloud, while the TonyPi platform handles sensing, communication, motion control, and feedback. A standardized JSON protocol connects the perception, reasoning, and control components.
\end{itemize}

\section{Related Work}
\subsection{Gesture Detection via Deep Learning}
The YOLO series has become a widely used foundation for real-time gesture detection because of its favorable balance between accuracy and efficiency \cite{ref1,ref2,ref3,ref4}. Two-stage detectors such as Faster R-CNN \cite{ref9} and Mask R-CNN \cite{ref14} often provide high accuracy but can be computationally expensive for edge devices. Achieving strong generalization, high localization accuracy, and real-time performance under illumination changes, background noise, and diverse gesture poses remains an active research problem.

A variety of bounding-box regression objectives have been proposed, including focal loss \cite{ref11}, IoU loss \cite{ref24}, generalized IoU (GIoU) \cite{ref20}, complete IoU (CIoU) and distance IoU (DIoU) \cite{ref21}, and SIoU \cite{ref22}. Attention mechanisms such as CBAM \cite{ref15}, squeeze-and-excitation modules \cite{ref16}, non-local networks \cite{ref18}, and coordinate attention \cite{ref19} have also been used to emphasize informative spatial regions and feature channels. Meanwhile, ResNet \cite{ref48}, VGG \cite{ref54}, Inception \cite{ref55}, Vision Transformer \cite{ref38}, Swin Transformer \cite{ref46}, and ConvNeXt \cite{ref47} have advanced feature extraction, whereas MobileNet \cite{ref49,ref50} and EfficientNet \cite{ref51} provide lightweight alternatives.

This work builds on YOLO11n and targets small or partially occluded gestures by embedding CBAM in the neck and emphasizing center-distance convergence through DIoU loss. The modifications introduce limited additional parameter overhead while preserving the one-stage detection pipeline.

\subsection{Multimodal Agents for Robotic Interaction}
Conventional robotic systems often use locally deployed, single-modal interfaces that restrict semantic interpretation and task planning in dynamic environments. VLMs and LLMs provide a more flexible basis for integrating visual and linguistic information. CLIP \cite{ref31}, BLIP \cite{ref32,ref33}, LLaVA \cite{ref34,ref35}, Flamingo \cite{ref36}, and MiniGPT-4 \cite{ref37} have demonstrated capabilities in visual question answering and multimodal content understanding. DINOv2 \cite{ref56} and masked autoencoders \cite{ref39} have further improved self-supervised visual representation learning.

In robotics, SayCan \cite{ref40}, VoxPoser \cite{ref41}, RT-2 \cite{ref42}, and CLIPort \cite{ref43} connect language-conditioned reasoning with task execution. Reasoning strategies such as chain-of-thought prompting \cite{ref29}, ReAct \cite{ref44}, and generative agents \cite{ref45} support decomposition and sequential decision-making. Nevertheless, fully local deployment of large multimodal models remains difficult on resource-limited robotic hardware. We therefore employ a cloud--edge dual-agent architecture in which the cloud performs model inference and semantic reasoning, while the robot retains low-latency sensing and motion control.

\subsection{Security Considerations for Cloud--Edge Interaction}
Cloud--edge robotic systems expose networked interfaces, remote inference services, and structured inter-component messages, making security validation relevant to reliable deployment. Penetration-testing workflows provide a systematic basis for identifying system-level attack surfaces and validating defensive controls \cite{ref57}. At the software-analysis level, recent smart-contract studies combine semantic representations, control-flow graphs, interface information, and interaction context to identify vulnerabilities \cite{ref58,ref60}, while chain-specific analyses emphasize that security risks and suitable tools vary across platforms \cite{ref59}. Cross-contract analysis further shows the value of examining complete interaction paths rather than isolated components \cite{ref61}. Although these studies target information systems and blockchain software rather than robotics, their emphasis on interface validation, interaction-aware analysis, and end-to-end attack paths motivates strict message-schema validation, command whitelisting, and cross-module consistency checks in the proposed framework.

\section{Method}
\subsection{System Architecture}
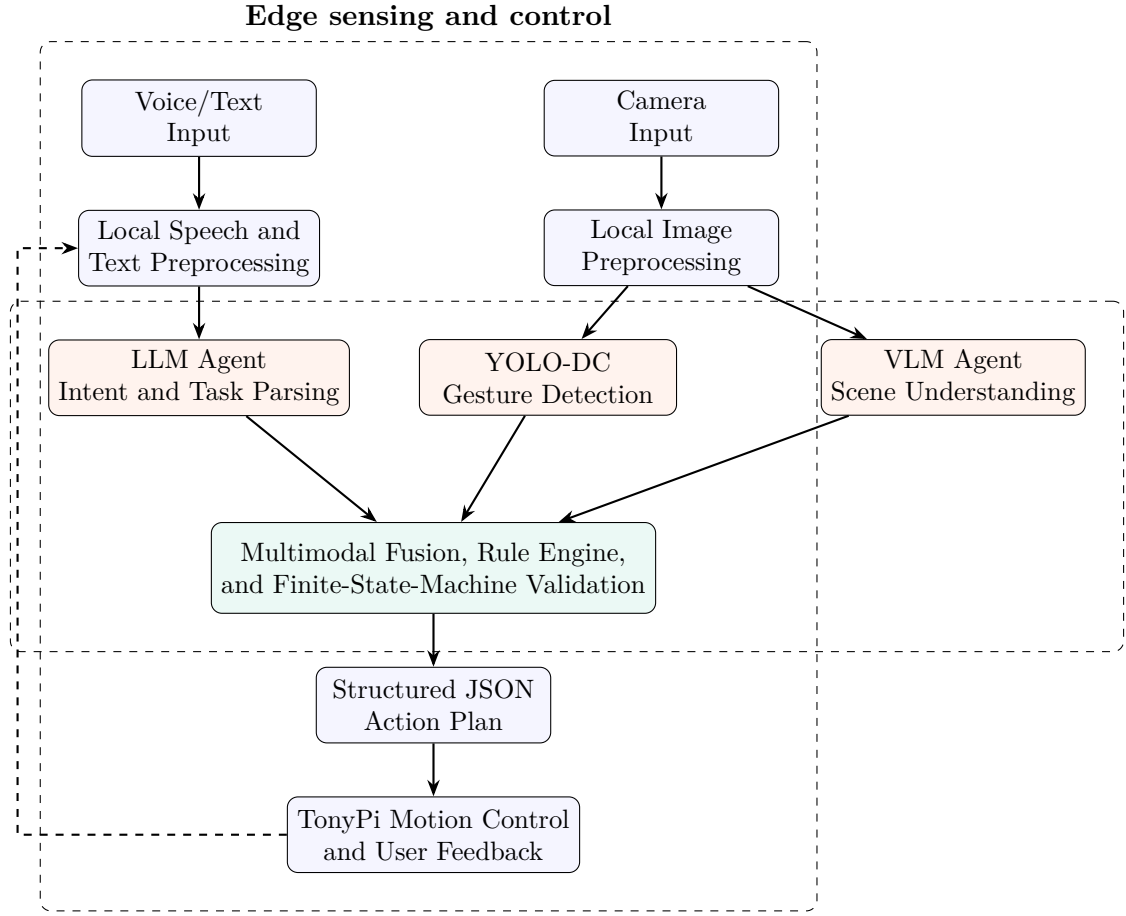
\begin{figure}[htbp]
\centering
\begin{tikzpicture}[
  font=\small,
  node distance=7mm and 10mm,
  >={Stealth[length=2.2mm]},
  block/.style={draw,rounded corners,align=center,minimum height=10mm,minimum width=31mm,fill=blue!4},
  cloud/.style={draw,rounded corners,align=center,minimum height=10mm,minimum width=34mm,fill=orange!8},
  fusion/.style={draw,rounded corners,align=center,minimum height=12mm,minimum width=42mm,fill=green!8},
  arrow/.style={->,thick}
]
\node[block] (voice) {Voice/Text\\Input};
\node[block, right=30mm of voice] (camera) {Camera\\Input};
\node[block, below=of voice] (speech) {Local Speech and\\Text Preprocessing};
\node[block, below=of camera] (image) {Local Image\\Preprocessing};
\node[cloud, below=of speech] (llm) {LLM Agent\\Intent and Task Parsing};
\node[cloud, below=of image, xshift=-15mm] (yolo) {YOLO-DC\\Gesture Detection};
\node[cloud, below right=7mm and 5.5mm of image] (vlm) {VLM Agent\\Scene Understanding};
\node[fusion, below=14mm of llm, xshift=31mm] (fusion) {Multimodal Fusion, Rule Engine,\\and Finite-State-Machine Validation};
\node[block, below=of fusion] (json) {Structured JSON\\Action Plan};
\node[block, below=of json] (control) {TonyPi Motion Control\\and User Feedback};

\draw[arrow] (voice) -- (speech);
\draw[arrow] (camera) -- (image);
\draw[arrow] (speech) -- (llm);
\draw[arrow] (image) -- (yolo);
\draw[arrow] (image) -- (vlm);
\draw[arrow] (llm) -- (fusion);
\draw[arrow] (yolo) -- (fusion);
\draw[arrow] (vlm) -- (fusion);
\draw[arrow] (fusion) -- (json);
\draw[arrow] (json) -- (control);
\draw[arrow,dashed] (control.west) -| ([xshift=-8mm]speech.west) |- (speech.west);

\node[draw,dashed,rounded corners,fit=(llm)(yolo)(vlm)(fusion),inner sep=5mm,label={[font=\bfseries]above:}] {};
\node[draw,dashed,rounded corners,fit=(voice)(camera)(speech)(image)(json)(control),inner sep=5mm,label={[font=\bfseries]above:Edge sensing and control}] {};
\end{tikzpicture}
\caption{Cloud--edge architecture of the proposed multimodal human--robot interaction framework.}
\label{fig:architecture}
\end{figure}

The proposed framework separates computationally demanding inference from real-time robot control. As shown in Figure~\ref{fig:architecture}, the local TonyPi platform captures voice and image inputs, compresses and transmits the data, receives a structured action plan, and executes the corresponding motion sequence. The cloud runs YOLO-DC for predefined gesture detection, a VLM agent for open-vocabulary scene understanding, and an LLM agent for intent interpretation and task decomposition. Their outputs are fused and checked by a rule engine and finite-state machines (FSMs) before being returned to the robot.

\subsection{YOLO-DC: Refined Gesture Detection}
\begin{figure}[htbp]
\centering
\includegraphics[width=0.95\linewidth]{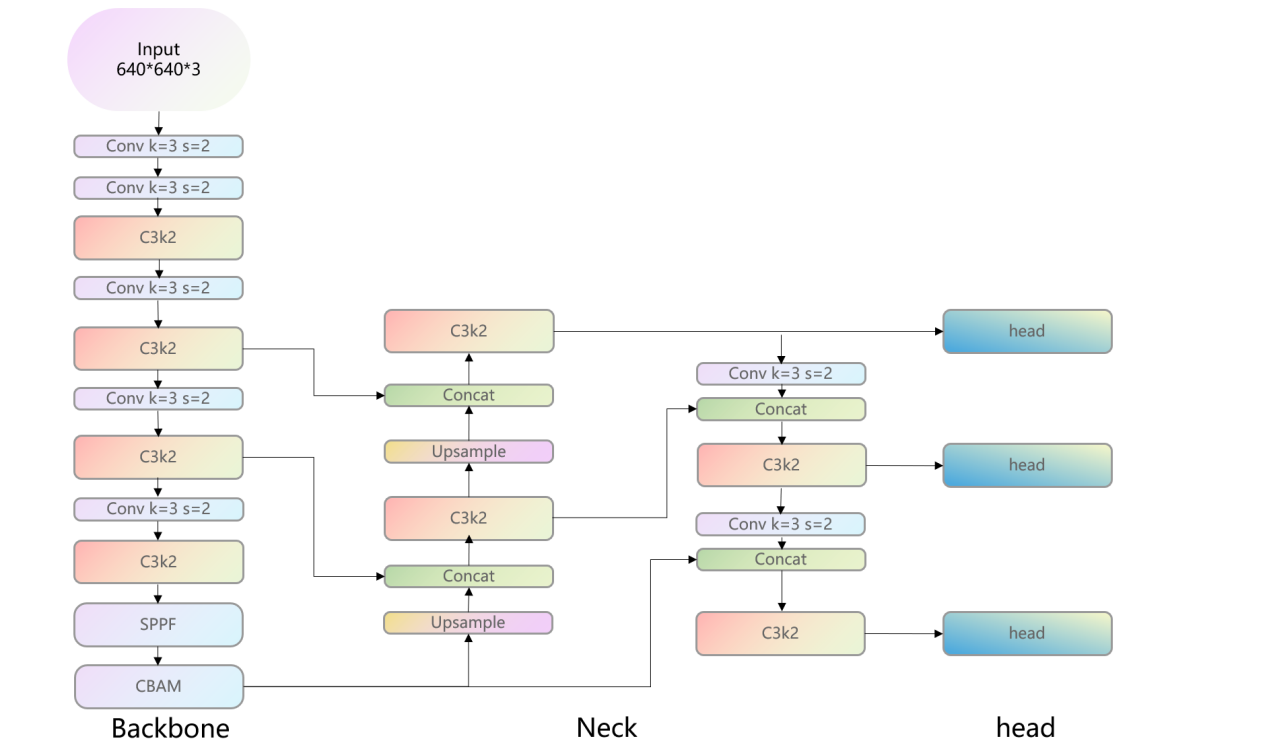}
\caption{Network architecture of YOLO-DC.}
\label{fig:yolo_dc}
\end{figure}

YOLO11n provides efficient one-stage detection, but small gestures and cluttered backgrounds can still cause weak feature responses and imprecise localization. YOLO-DC addresses these limitations by integrating CBAM \cite{ref15} into the neck and using DIoU loss \cite{ref21} for bounding-box regression. The network structure is shown in Figure~\ref{fig:yolo_dc}.

\subsubsection{CBAM Attention Mechanism}
CBAM sequentially applies channel and spatial attention to enhance informative feature responses. For an input feature map $F$, the operations are
\begin{align}
F' &= M_c(F) \otimes F, \label{eq:channel}\\
F'' &= M_s(F') \otimes F', \label{eq:spatial}
\end{align}
where $M_c$ and $M_s$ denote the channel and spatial-attention maps, respectively, $\otimes$ denotes element-wise multiplication, and $F''$ is the refined output feature map. The CBAM structure is illustrated in Figure~\ref{fig:cbam}.

\tikzset{
  every node/.style={font=\sffamily},
  flow/.style={-{Stealth[length=2.4mm,width=1.7mm]}, line width=0.72pt,
               draw=black!85, rounded corners=1.5mm},
  panel/.style={draw=panelblue, line width=0.85pt, rounded corners=7mm},
  module/.style={draw=orange, line width=0.85pt, rounded corners=3mm,
                 fill=white, align=center, inner sep=2.2mm},
  label/.style={font=\sffamily\small},
  title/.style={font=\sffamily\bfseries\large},
  sum/.style={circle, draw=black!80, fill=white, line width=0.75pt,
              minimum size=6.2mm, inner sep=0pt, font=\sffamily\large},
}

\newcommand{\featurecuboid}[7]{%
  \path[draw=black!90, fill=frontblue, line width=0.85pt]
    (#1,#2) rectangle ++(#3,#4);
  \path[draw=black!90, fill=topblue, line width=0.85pt]
    (#1,#2+#4) -- ++(0.46,0.42) -- ++(#3,0) -- ++(-0.46,-0.42) -- cycle;
  \path[draw=black!90, fill=sideblue, line width=0.85pt]
    (#1+#3,#2) -- ++(0.46,0.42) -- ++(0,#4) -- ++(-0.46,-0.42) -- cycle;
  \node[font=\sffamily\small, align=center, anchor=north]
    at (#1+#3/2+#6,#2-0.10+#7) {#5};
}

\newcommand{\hstrip}[5]{%
  \path[draw=black!90, fill=#5, line width=0.75pt]
    (#1,#2) rectangle ++(#3,#4);
  \path[draw=black!90, fill=#5!72!black, line width=0.75pt]
    (#1,#2+#4) -- ++(0.09,0.09) -- ++(#3,0) -- ++(-0.09,-0.09) -- cycle;
  \path[draw=black!90, fill=#5!62!black, line width=0.75pt]
    (#1+#3,#2) -- ++(0.09,0.09) -- ++(0,#4) -- ++(-0.09,-0.09) -- cycle;
}

\newcommand{\vplane}[7]{%
  \path[draw=black!90, fill=#5, line width=0.78pt]
    (#1,#2) -- (#1+#3,#2+0.25) -- (#1+#3,#2+#4+0.25) -- (#1,#2+#4) -- cycle;
  \path[draw=black!90, fill=#6, line width=0.78pt]
    (#1,#2+#4) -- (#1+#3,#2+#4+0.25) -- (#1+#3+0.13,#2+#4+0.36)
    -- (#1+0.13,#2+#4+0.11) -- cycle;
  \path[draw=black!90, fill=#7, line width=0.78pt]
    (#1+#3,#2+0.25) -- (#1+#3+0.13,#2+0.36) -- (#1+#3+0.13,#2+#4+0.36)
    -- (#1+#3,#2+#4+0.25) -- cycle;
}

\begin{figure}[htbp]
    \centering 
    \begin{tikzpicture}[x=0.8cm,y=0.8cm]
      \path[use as bounding box] (-0.03,-0.02) rectangle (18.27,11.63);
    
      \draw[panel] (0.05,8.05) rectangle (18.22,11.55);
      \node[title, anchor=north east] at (17.82,11.60) {Channel Attention Module};
    
      \featurecuboid{0.72}{8.82}{1.66}{1.16}{Input feature}{0.2}{0}
    
      \node[label] (maxlabel) at (4.72,11.08) {Maxpool};
      \hstrip{4.02}{10.55}{1.74}{0.20}{darkorange}
      \node[label] (avglabel) at (4.72,9.02) {Avgpool};
      \hstrip{4.20}{8.43}{1.72}{0.20}{lightgreen}
    
      \draw[flow] (3.02,9.94) to[out=25,in=180] (3.96,10.67);
      \draw[flow] (3.02,9.45) to[out=-25,in=180] (4.14,8.56);
    
      \path[draw=black!90, fill=white, line width=0.75pt]
        (7.05,8.85) rectangle (7.34,10.82);
      \path[draw=black!75, fill=black!6, line width=0.65pt]
        (7.34,8.92) -- (7.46,9.00) -- (7.46,10.90) -- (7.34,10.82) -- cycle;
      \path[draw=black!90, fill=white, line width=0.75pt]
        (7.57,9.28) rectangle (7.84,10.58);
      \path[draw=black!75, fill=black!6, line width=0.65pt]
        (7.84,9.35) -- (7.96,9.43) -- (7.96,10.66) -- (7.84,10.58) -- cycle;
      \path[draw=black!90, fill=white, line width=0.75pt]
        (8.10,8.85) rectangle (8.39,10.82);
      \path[draw=black!75, fill=black!6, line width=0.65pt]
        (8.39,8.92) -- (8.51,9.00) -- (8.51,10.90) -- (8.39,10.82) -- cycle;
      \node[font=\sffamily\bfseries\small] at (7.77,8.46) {MLP};
    
      \draw[flow] (5.86,10.64) to[out=-15,in=160] (6.88,9.85);
      \draw[flow] (5.98,8.54) to[out=20,in=200] (6.88,9.42);
    
      \hstrip{10.28}{10.54}{1.95}{0.20}{orange}
      \hstrip{10.28}{8.48}{1.95}{0.20}{green}
      \draw[flow] (8.66,9.90) to[out=15,in=180] (10.20,10.66);
      \draw[flow] (8.66,9.46) to[out=-15,in=180] (10.20,8.60);
    
      \node[sum] (sumtop) at (14.02,9.68) {$+$};
      \draw[flow] (12.43,10.65) to[out=-10,in=85] (sumtop.north);
      \draw[flow] (12.43,8.58) to[out=5,in=275] (sumtop.south);
        
      \hstrip{15.90}{9.47}{1.95}{0.20}{yellowatt}
      \draw[flow] (sumtop.east) -- (15.79,9.65);
      \node[font=\sffamily\bfseries\small, align=center] at (16.83,8.86)
          {Channel\\Attention};

      \draw[panel] (0.05,4.02) rectangle (18.22,7.66);
      \node[title, anchor=north east] at (17.82,7.45) {Spatial Attention Module};
    
      \featurecuboid{0.95}{4.90}{1.66}{1.16}{Channel-refined feature}{0.7}{0}
    
      \node[label] at (5.18,7.43) {Maxpool};
      \vplane{5.07}{6.08}{0.22}{0.81}{poolblue}{poolblue!72!white}{poolblue!65!black}
      \node[label] at (5.18,5.72) {Avgpool};
      \vplane{5.07}{4.46}{0.22}{0.81}{poolgreen}{poolgreen!72!white}{poolgreen!65!black}
    
      \draw[flow] (3.14,6.08) to[out=18,in=180] (4.92,6.56);
      \draw[flow] (3.14,5.47) to[out=-18,in=180] (4.92,4.86);
    
      \vplane{9.14}{4.60}{0.20}{2.40}{convside}{convtop}{convfront!70!black}
      \node[label] at (9.40,4.31) {$7\!\times\!7$ conv layer};
      \draw[flow] (5.46,6.53) to[out=5,in=165] (8.63,5.91);
      \draw[flow] (5.46,4.83) to[out=-4,in=195] (8.63,5.67);
    
      \draw[flow] (10.43,5.94) -- (16.02,5.94);
      \node[label, fill=white, inner sep=1.2pt] at (12.63,6.25) {Sigmoid};
      \vplane{16.29}{4.58}{0.22}{1.90}{spatialfront}{spatialtop}{spatialside}
      \node[label] at (16.0,4.31) {Spatial Attention};
    
      \draw[panel] (0.05,0.07) rectangle (18.22,3.72);
      \node[title, anchor=north east] at (17.82,3.50) {CBAM Module};
    
      \featurecuboid{0.72}{0.92}{1.66}{1.16}{Input feature}{0}{0}
      \node[module, minimum width=2.35cm, minimum height=1.30cm] (cam)
        at (5.76,2.53) {Channel\\Attention\\Module};
      \node[sum] (sumone) at (8.17,1.39) {$\times$};
    
      \node[module, minimum width=2.35cm, minimum height=1.30cm] (sam)
        at (11.30,2.53) {Spatial\\Attention\\Module};
      \node[sum] (sumtwo) at (13.74,1.39) {$\times$};
    
      \draw[flow] (2.86,1.39) -- (sumone.west);
      \draw[flow] (sumone.east) -- (sumtwo.west);
      \draw[flow] (sumtwo.east) -- (15.66,1.39);
    
      \draw[flow] (2.86,2.05) to[out=25,in=180] (cam.west);
      \draw[flow] (cam.east) to[out=-5,in=90] (sumone.north);
      \draw[flow] (sumone.north) to[out=45,in=180] (sam.west);
      \draw[flow] (sam.east) to[out=-5,in=90] (sumtwo.north);
    
      \featurecuboid{15.82}{0.82}{1.66}{1.16}{Refined feature}{-0.3}{0}
    \end{tikzpicture}
  \caption{The Architecture of CBAM} 
  
  \label{fig:cbam}
\end{figure}
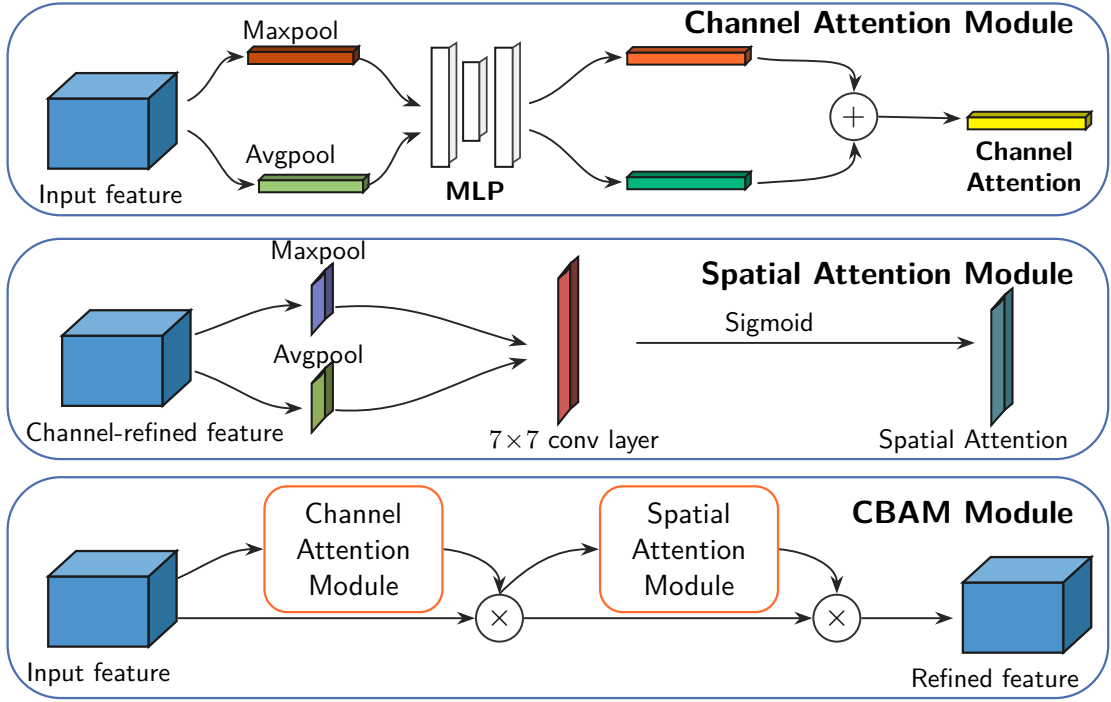

Although the C2PSA module in YOLO11n already provides spatial attention, its feature aggregation may remain insufficient for small gestures. We therefore place CBAM after the C2PSA stage in the neck. Channel attention emphasizes feature channels associated with gesture cues, and spatial attention highlights target regions. According to the model summary, the modification adds approximately $0.1$ million parameters.

\subsubsection{DIoU-Based Bounding-Box Regression}
CIoU considers overlap, center-point distance, and aspect-ratio consistency. In this study, DIoU is used to place greater emphasis on center-distance convergence and to simplify the geometric penalty term. The DIoU loss is
\begin{equation}
\mathcal{L}_{\mathrm{DIoU}}
= 1-\mathrm{IoU}
+ \frac{\rho^2(\mathbf{b},\mathbf{b}^{\mathrm{gt}})}{c^2},
\label{eq:diou}
\end{equation}
where $\mathrm{IoU}$ is the intersection-over-union ratio, $\mathbf{b}$ and $\mathbf{b}^{\mathrm{gt}}$ are the center points of the predicted and ground-truth boxes, respectively, $\rho(\cdot,\cdot)$ is the Euclidean distance, and $c$ is the diagonal length of the smallest enclosing box. The center-distance penalty remains informative even when the two boxes do not overlap and encourages direct convergence of their centers.

\subsection{Cloud--Edge Collaborative Task Flow}
\subsubsection{Local Perception and Transmission}
A voice command initiates the interaction process. The local perception layer captures the audio signal, digitizes it in pulse-code modulation format, applies noise reduction, and performs speech-to-text conversion. Simultaneously, the camera captures image frames, which are denoised, resized, and JPEG-encoded. A JPEG quality factor of 85 is used to reduce communication overhead. In the reported test, this setting reduced the image size from approximately $3.2$~MB to $1.9$~MB while retaining sufficient visual detail for gesture detection and scene understanding. The processed text and image data are then transmitted to the cloud.

\subsubsection{Cloud Fusion and Action Planning}
In the cloud, YOLO-DC detects predefined hand gestures, while the VLM agent performs scene understanding and visual question answering. The LLM agent fuses the voice command, gesture results, and visual description to infer user intent and generate a candidate action sequence. For example, a query such as ``What is in front of you?'' produces a natural-language response grounded in the visual analysis. A behavior command such as \texttt{celebrate} produces a structured action token such as \texttt{celebrate()} together with a spoken response.

A rule engine verifies action compatibility, and FSMs enforce execution order and state consistency. Cross-agent communication uses a unified JSON schema containing the request type, recognized intent, object classes, colors, positions, confidence scores, response text, and action sequence. The local decision layer parses this message, triggers text-to-speech feedback when required, and invokes the corresponding motion-control interface.

These mechanisms provide functional validation at the message and command levels, but they do not constitute a complete cybersecurity assessment. A production deployment should additionally implement authenticated and encrypted communication, access control, rate limiting, audit logging, and systematic penetration testing \cite{ref57}.

\section{Experiments}
\subsection{Experimental Setup}
\subsubsection{Datasets}
\begin{figure}[htbp]
\centering
\includegraphics[width=0.92\linewidth]{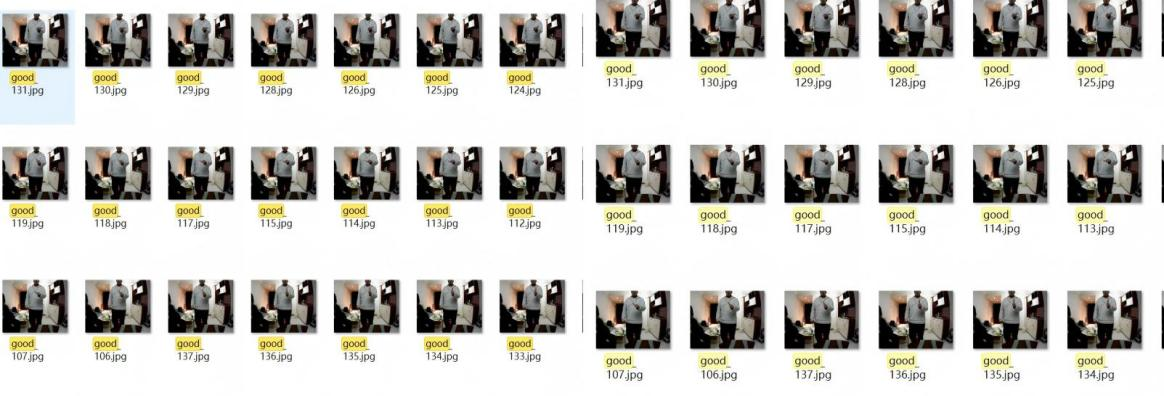}
\caption{Representative samples from the custom dataset.}
\label{fig:custom_samples}
\end{figure}

\begin{figure}[htbp]
\centering
\includegraphics[width=0.92\linewidth]{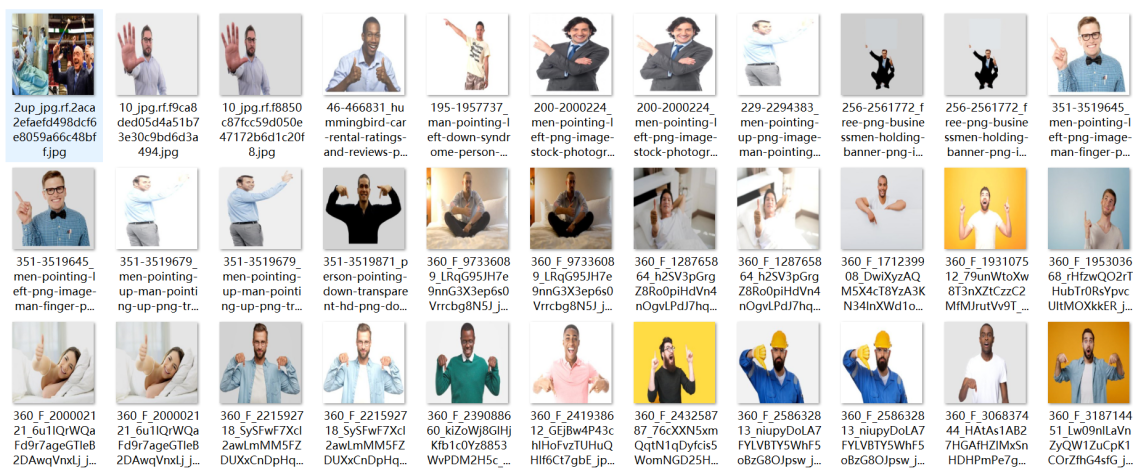}
\caption{Representative samples from the public dataset.}
\label{fig:public_samples}
\end{figure}

Two datasets are used for gesture detection:
\begin{itemize}
  \item \textbf{Public dataset:} a publicly available gesture dataset containing 600 images from six classes. Representative samples are shown in Figure~\ref{fig:public_samples}.
  \item \textbf{Custom dataset:} a dataset collected to approximate practical robot-interaction conditions. It contains three classes: \emph{good}, \emph{left}, and \emph{palm}. Videos are recorded with a Raspberry Pi camera, and one frame is extracted every five frames. Representative samples are shown in Figure~\ref{fig:custom_samples}.
\end{itemize}

For each dataset, the images are divided into training and validation subsets at a ratio of 2:1. The validation subset is used for model selection and for reporting the results in this paper.

\subsubsection{Implementation Details}
Experiments are conducted using an NVIDIA GeForce RTX 4090 GPU and PyTorch 2.1.0. The models are trained for 300 epochs with stochastic gradient descent, a momentum of 0.937, a batch size of 32, and an initial learning rate of 0.01. Input images are resized to $640\times640$ pixels. A dropout rate of 0.5 is used in the training configuration to mitigate overfitting.

\subsection{Gesture Detection Results}
\subsubsection{Comparison with Mainstream Models}
YOLO-DC is compared with YOLOv5n, YOLOv8n, and the baseline YOLO11n on both datasets. As reported in Table~\ref{tab:comparison}, YOLO-DC achieves the best performance across all reported metrics. On the public dataset, it reaches a precision of 98.9\% and an mAP@0.5 of 90.7\%. On the custom dataset, it achieves a precision of 95.0\% and an mAP@0.5 of 92.7\%. Relative to YOLO11n on the custom dataset, the precision increases by 16.0 percentage points and the mAP@0.5 increases by 6.0 percentage points.

\begin{table}[htbp]
\centering
\caption{Model performance on the public and custom datasets.}
\label{tab:comparison}
\small
\setlength{\tabcolsep}{7pt}
\begin{tabular}{lcccc}
\toprule
\multirow{2}{*}{Model} & \multicolumn{2}{c}{Public dataset} & \multicolumn{2}{c}{Custom dataset}\\
\cmidrule(lr){2-3}\cmidrule(lr){4-5}
& Precision (\%) & mAP@0.5 (\%) & Precision (\%) & mAP@0.5 (\%)\\
\midrule
YOLOv5n          & 92.8 & 84.5 & 69.0 & 82.9\\
YOLOv8n          & 97.3 & 88.4 & 72.3 & 84.6\\
YOLO11n          & 97.5 & 88.4 & 79.0 & 86.7\\
YOLO-DC (ours)   & \textbf{98.9} & \textbf{90.7} & \textbf{95.0} & \textbf{92.7}\\
\bottomrule
\end{tabular}
\end{table}

\begin{figure}[htbp]
\centering
\includegraphics[width=0.94\linewidth]{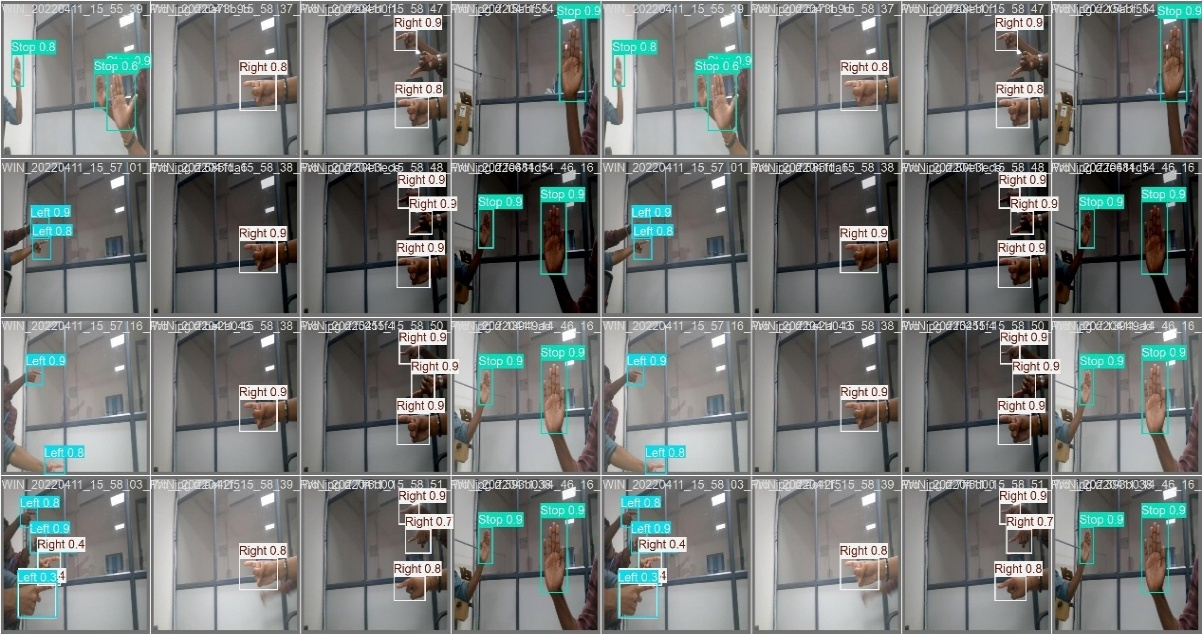}
\caption{Comparison of detection results on the public dataset: YOLO-DC (left) and YOLO11n (right).}
\label{fig:public_detection}
\end{figure}

\begin{figure}[htbp]
\centering
\includegraphics[width=0.94\linewidth]{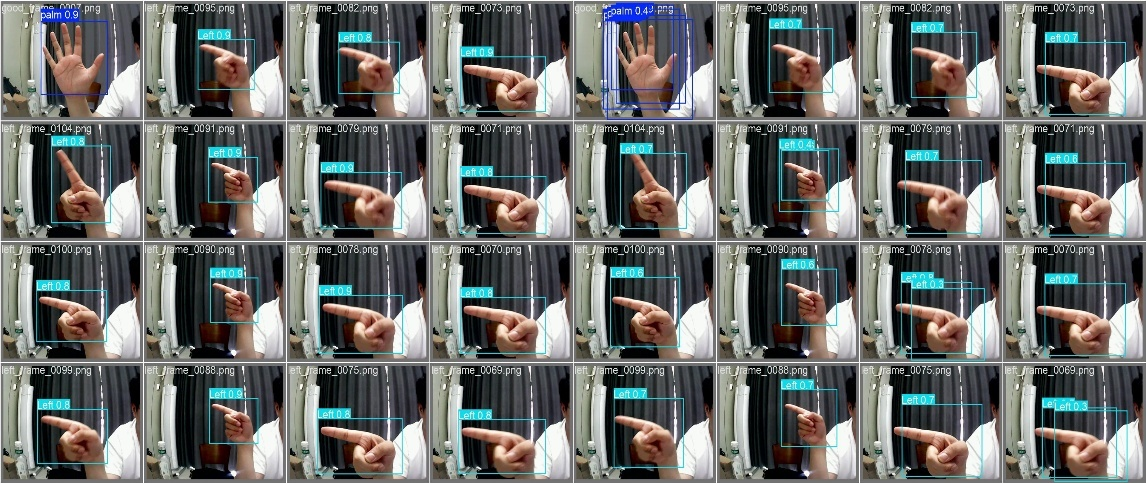}
\caption{Comparison of detection results on the custom dataset: YOLO-DC (left) and YOLO11n (right).}
\label{fig:custom_detection}
\end{figure}

Visual comparisons on the public and custom datasets are shown in Figure~\ref{fig:public_detection} and \ref{fig:custom_detection}, respectively. Compared with YOLO11n, YOLO-DC generally provides more precise bounding boxes and higher confidence scores in cluttered or partially occluded scenes.

\subsubsection{Ablation Study}
An ablation study is conducted on the public dataset to evaluate the contributions of DIoU and CBAM. As shown in Table~\ref{tab:ablation}, replacing the baseline regression objective with DIoU increases recall from 94.0\% to 95.0\% and mAP@0.5 from 88.4\% to 89.6\%. Adding CBAM alone increases mAP@0.5 from 88.4\% to 88.9\%, although precision and recall decrease slightly. Combining both modifications produces the highest precision, recall, mAP@0.5, and calculated $F_1$ score. The $F_1$ values in Table~\ref{tab:ablation} are calculated from the reported precision and recall as $F_1 = 2PR/(P + R)$.

\begin{table}[htbp]
\centering
\caption{Ablation study on the public dataset.}
\label{tab:ablation}
\small
\setlength{\tabcolsep}{8pt}
\begin{tabular}{lcccc}
\toprule
Model & Precision (\%) & Recall (\%) & mAP@0.5 (\%) & $F_1$ (\%)\\
\midrule
YOLO11n         & 97.5 & 94.0 & 88.4 & 95.7\\
YOLO11n $+$ DIoU  & 97.4 & 95.0 & 89.6 & 96.2\\
YOLO11n $+$ CBAM  & 96.4 & 94.0 & 88.9 & 95.2\\
YOLO-DC (ours)  & \textbf{98.9} & \textbf{95.0} & \textbf{90.7} & \textbf{96.9}\\
\bottomrule
\end{tabular}
\end{table}

\begin{figure}[htbp]
\centering
\includegraphics[width=0.98\linewidth]{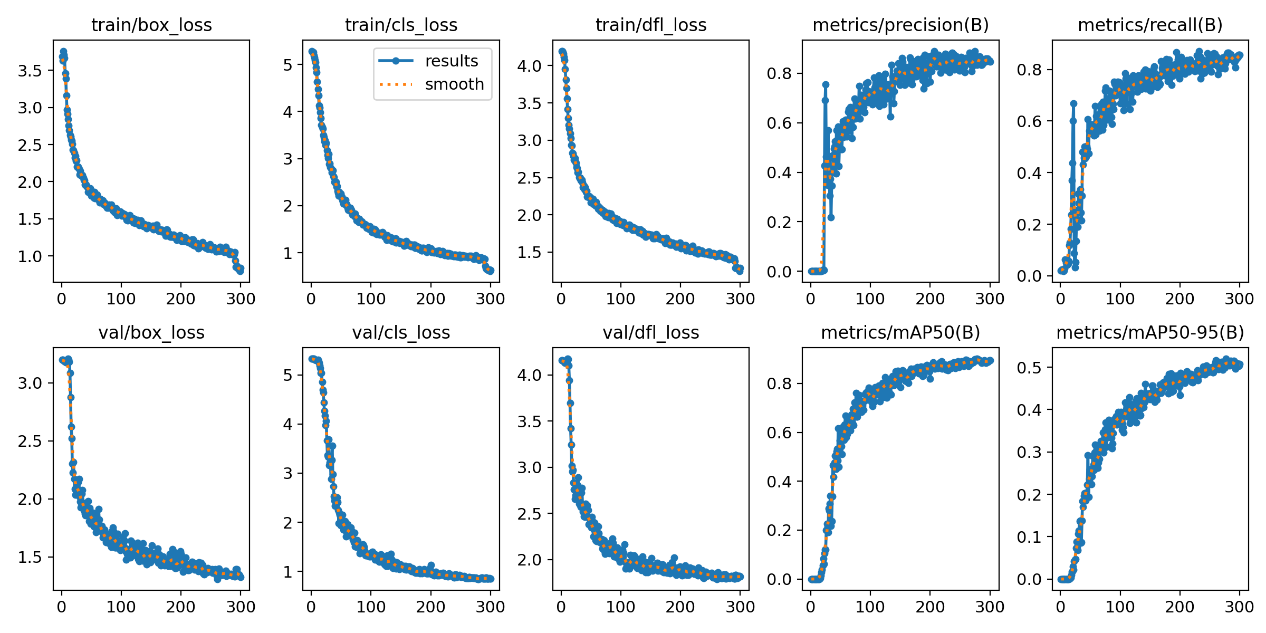}
\caption{Training and validation curves of YOLO-DC on the public dataset.}
\label{fig:training}
\end{figure}

The training curves in Figure~\ref{fig:training} show stable convergence of the box-regression, classification, and distribution focal losses. Precision, recall, and mAP increase progressively and stabilize during training.

\subsection{System-Level Evaluation}
The multimodal interaction system is evaluated on the TonyPi platform in terms of task-execution success, response diversity, and user satisfaction.

\subsubsection{Task-Execution Success Rate}
\begin{figure}[htbp]
\centering
\begin{tikzpicture}[
  font=\small,
  panel/.style={draw=black!60, rounded corners=2pt, line width=0.5pt,
                inner sep=6pt, text width=0.88\linewidth, align=left},
  tag/.style={font=\bfseries\footnotesize, text=black!75},
  arrow/.style={-{Latex[length=2.2mm]}, line width=0.6pt, draw=black!65}
]
\node[panel, fill=black!3] (input) {
  \textbf{User query}\hfill\texttt{request\_type: scene\_query}\\[2pt]
  ``What is in front of you?''
};

\node[panel, fill=black!1, below=5mm of input] (vlm) {
  \textbf{VLM scene description}\\[2pt]
  A laptop, a keyboard, a smartphone, and a small speaker are visible on the desk.
};

\node[panel, fill=black!3, below=5mm of vlm] (output) {
  \textbf{Multimodal-agent output}\\[2pt]
  \texttt{intent: describe\_scene}\\
  \texttt{objects: [laptop, keyboard, smartphone, speaker]}\\
  \texttt{action\_sequence: []}\\
  \texttt{response: ``A laptop, a keyboard, a smartphone, and a small speaker are in front of me.''}
};

\draw[arrow] (input) -- node[right,tag]{visual-language inference} (vlm);
\draw[arrow] (vlm) -- node[right,tag]{fusion and validation} (output);
\end{tikzpicture}
\caption{English-language example of the scene-understanding and structured-response output generated by the multimodal agent.}
\label{fig:scene_output}
\end{figure}
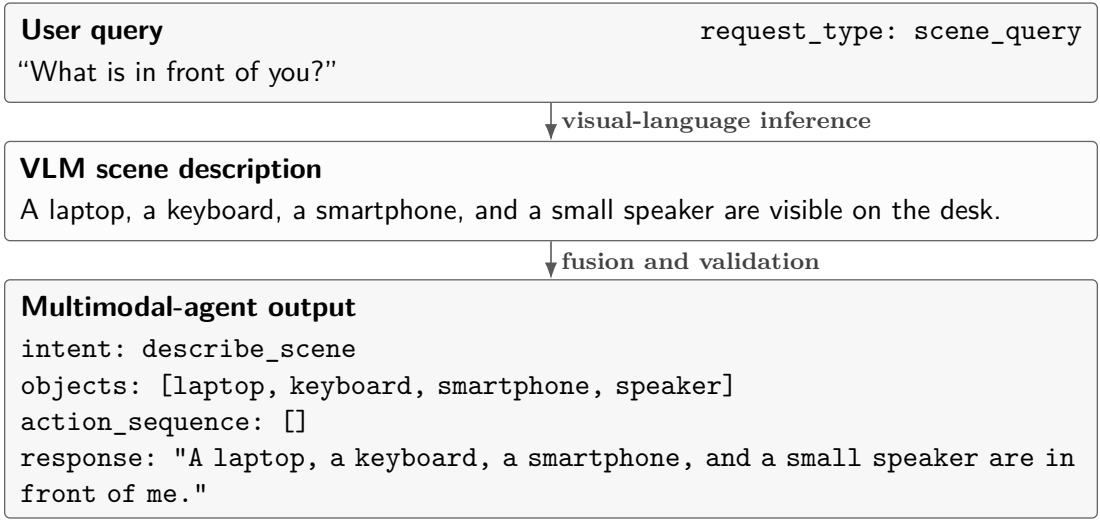

Three command categories are evaluated: single-action, composite-action, and vision-dependent tasks. Table~\ref{tab:task_success} shows that single-action commands achieve a success rate of 95\%, composite-action commands achieve 88\%, and vision-dependent tasks achieve 82\%. The rule engine rejects 90\% of the tested invalid action combinations, such as incompatible consecutive motion commands. Failures in vision-dependent tasks are primarily associated with visual-interpretation errors, communication delays, and response-format inconsistencies. The measured mean VLM response latency is approximately 210~ms. Figure~\ref{fig:scene_output} presents an example of the multimodal agent's scene-understanding output.

\begin{table}[htbp]
\centering
\caption{Task-execution success rates.}
\label{tab:task_success}
\begin{tabular}{lc}
\toprule
Task category & Success rate (\%)\\
\midrule
Single-action command & 95\\
Composite-action command & 88\\
Vision-dependent task & 82\\
\bottomrule
\end{tabular}
\end{table}

\subsubsection{Feedback Diversity and User Satisfaction}
To assess response diversity, 500 responses are generated from 50 predefined templates. The resulting repetition rate is below 5\%. In addition, 30 participants rate four predefined interaction scenarios---Wave, Kick Ball, Twist, and Celebrate---on a five-point scale. Table~\ref{tab:satisfaction} reports the rating distributions. The mean scores calculated from these distributions are 3.83, 3.67, 3.70, and 3.57, respectively, yielding an overall mean of 3.69 out of 5.

\begin{table}[h]
\centering
\caption{User-satisfaction rating distribution.}
\label{tab:satisfaction}
\small
\begin{tabular}{lcccc}
\toprule
Rating & Wave & Kick Ball & Twist & Celebrate\\
\midrule
5 points & 12 & 13 & 12 & 12\\
4 points & 8  & 5  & 6  & 5\\
3 points & 6  & 4  & 4  & 5\\
2 points & 1  & 5  & 7  & 4\\
1 point  & 3  & 3  & 1  & 4\\
\midrule
Mean score & 3.83 & 3.67 & 3.70 & 3.57\\
\bottomrule
\end{tabular}
\end{table}

\section{Conclusion and Future Work}
This paper presents a cloud--edge multimodal human--robot interaction framework that combines YOLO-DC gesture detection with coordinated LLM and VLM agents. YOLO-DC integrates CBAM into the neck of YOLO11n and uses DIoU loss to improve localization and feature discrimination for small or partially occluded gestures. On the public dataset, the model achieves a precision of 98.9\% and an mAP@0.5 of 90.7\%, on the custom dataset, it achieves 95.0\% precision and 92.7\% mAP@0.5.
The cloud--edge architecture separates computationally intensive perception and reasoning from local sensing and control. System-level evaluation produces success rates of 95\%, 88\%, and 82\% for single-action, composite-action, and vision-dependent tasks, respectively. The participant study yields an overall mean satisfaction score of 3.69 out of 5. These results support the practical feasibility of combining refined gesture detection, multimodal scene understanding, and structured action planning on resource-constrained robotic platforms.

Future work will focus on model compression, pruning, quantization, and knowledge distillation to reduce cloud dependence and enable more inference on edge hardware. Additional sensory modalities, including depth and tactile information, will be incorporated to improve robustness under occlusion and rapid environmental change. Larger-scale evaluations in industrial inspection, smart-home assistance, education, and healthcare scenarios will also be conducted to examine generalization and long-term usability.

\section*{Acknowledgments}
The authors express deepest gratitude to Xia Ran. AI-based tools were used for language polishing during manuscript preparation.

\end{document}